\documentclass[final,leqno]{article}
\usepackage[T1]{fontenc} 
\usepackage{ae,aecompl}	 
\usepackage{times}
\usepackage{soul,fancyhdr}
\usepackage{amsfonts}\usepackage{amsmath}\usepackage{amssymb}
\usepackage[hang,small,bf]{caption}
\usepackage[latin1]{inputenc} 
\usepackage{hyperref}\usepackage{breakurl} 
\usepackage{graphicx} 
\usepackage{float,booktabs}
\usepackage{subfig}
\usepackage[usenames,dvipsnames]{color}

\begin{document}

\title{Climate-driven statistical models as effective predictors of local dengue incidence in Costa Rica: A Generalized Additive Model and Random Forest approach
}

\maketitle

\begin{center}
\author{Paola V\'asquez\footnote{Corresponding author: \url{paola.vasquez@ucr.ac.cr}\\
Present address: Escuela de Salud P\'ublica,Universidad de Costa Rica. San Pedro de Montes de Oca, San Jos\'e, Costa Rica, 11501.},
Antonio Lor\'ia\footnote{Centro de Investigaci\'on en Matem\'atica Pura y Aplicada (CIMPA), Escuela de Matem\'atica, Universidad de Costa Rica. San Pedro de Montes de Oca, San Jos\'e, Costa Rica, 11501. Email: \url{antonio.loria@ucr.ac.cr}},
Fabio Sanchez\footnote{Centro de Investigaci\'on en Matem\'atica Pura y Aplicada (CIMPA), Escuela de Matem\'atica, Universidad de Costa Rica. San Pedro de Montes de Oca, San Jos\'e, Costa Rica, 11501. Email: \url{fabio.sanchez@ucr.ac.cr}} and Luis A. Barboza\footnote{Centro de Investigaci\'on en Matem\'atica Pura y Aplicada (CIMPA), Escuela de Matem\'atica, Universidad de Costa Rica. San Pedro de Montes de Oca, San Jos\'e, Costa Rica, 11501Email: \url{luisalberto.barboza@ucr.ac.cr}}}
\end{center}


\begin{abstract}
Climate has been an important factor in shaping the distribution and incidence of dengue cases in tropical and subtropical countries. In Costa Rica, a tropical country with distinctive micro-climates, dengue has been endemic since its introduction in 1993, inflicting substantial economic, social, and public health repercussions. Using the number of dengue reported cases and climate data from 2007-2017, we fitted a prediction model applying a Generalized Additive Model (GAM) and Random Forest (RF) approach, which allowed us to retrospectively predict the relative risk of dengue in five climatological diverse municipalities around the country.
\end{abstract}



\section{Introduction}
Dengue fever is a mosquito-borne viral infection of global significance. Currently, more than 120 tropical and subtropical countries in Africa, the Americas, and the Asia Pacific regions report endemic circulation of the dengue viruses (DENV) and their main mosquito vectors: \textit{Aedes aegypti} and \textit{Aedes albopictus} \cite{brady2012,bhatt2013} where they cause seasonal epidemics that disrupt the health and well being of the population and inflict substantial socioeconomic impact to households, health-care systems, and governments \cite{gubler2012economic,castro2017disease}.  

In Costa Rica, as in most of the Americas, the reintroduction and dissemination of \textit{Aedes aegypti} took place during the 1970s \cite{dick2012history,paho}. However, it was until September 1993 that the first dengue cases were reported on the Pacific coast \cite{minsa} when autochthonous transmission of DENV-1 was confirmed \cite{morice2010dengue}. Since then, three of the four serotypes of the virus (DENV-1, DENV-2, DENV-3) have circulated the national territory, with peaks of transmission that exhibit both seasonal and inter-annual variability \cite{minsa}. Over 370,000 suspected and confirmed cases have been reported by the Ministry of Health \cite{minsa}, of which, more than 45,000 have required hospital care \cite{ccss}.

With the high burden that DENV infections represent to the country, where, as in most endemic regions, traditional control measures have proven ineffective to sustain long-term trends in cases-reduction \cite{falcon2019dengue}, surveillance, prevention, and control of dengue is a public health challenge that requires specific and cost-effective strategies \cite{world2012global}. In this effort, and as a worldwide strategy for reducing dengue incidence, the World Health Organization (WHO) is highlighting the importance of determining sensitive indicators for dengue outbreaks as early warning signals \cite{world2012global}, in which climate and weather variables have shown to play an essential role \cite{hopp2003worldwide,descloux2012climate,ebi2016dengue}. Specifically, variables such as temperature, precipitation, humidity and El Ni\~no Southern Oscillation (ENSO), have been closely correlated to the occurrence of dengue cases and the seasonality of dengue epidemics \cite{chowell2006climate,carvajal2018machine,wu2007weather}. 

Changes in these climate conditions influence the ecology of the DENV by modulating vector mosquito population dynamics, viral replication, and transmission, as well as, human behavior \cite{brady2014,morin2013}. It has been observed that transmission of DENV occurs between 18$^{\circ}$C - 34$^{\circ}$C, with maximal transmission peaks in the range of 26$^{\circ}$C - 29$^{\circ}$C \cite{mordecai2017detecting}. At higher temperatures, the duration of the life cycle decreases \cite{yang2009,tun2000effects}, biting activity increases \cite{watts1987,rueda1990temperature,ebi2016dengue} and the extrinsic incubation period becomes shorter \cite{chan2012,xiao2014effect}, prolonging the infective days of the mosquito \cite{ebi2016dengue}. Precipitation provides habitat for the aquatic stages of the life cycle and influences vector distribution \cite{morin2013}. Moreover, heavy rainfall events can decrease mosquito abundance by flushing larvae from containers \cite{koenraadt2008flushing,benedum2018statistical}, and drought events can increase household water containers \cite{valdez2018impact}. Humidity also affects the biology of the mosquito as low levels of humidity have been associated with lower levels of oviposition \cite{costa2010} and a decreased survival rate \cite{canyon1999}. Other studies have also associated ENSO with dengue occurrence, as El Ni\~no and La Ni\~na events are associated with an increased probability of droughts in some areas and excess of rainfall in other regions \cite{gagnon2001dengue,tipayamongkholgul2009effects,fuller2009,hales1999nino}.  

The influence that these variables have on dengue transmission, and their potential use in the decision-making process, have prompted the use of numerous statistical models \cite{rodrigues2016temporal,lowe2017climate,lowe2018nonlinear}, which have shown promising results for the development and implementation of predictive models. Among them, Generalized Additive Models (GAM) and the Random Forest method (RF), have previously proven to be valuable tools for time series prediction analysis \cite{xiao2018weather,carvajal2018machine,hii2012forecast}. However, results vary among studies, as the complex role of local immunity patterns, public health interventions, population structure, and mobility, means that the relationship between dengue incidence and climate variables often differs across locations \cite{naish2014climate}. 

Given the weekly dengue data and climate information provided by the Ministry of Health and National Meteorological Institute, we analyzed the influence of temperature, precipitation, relative humidity and ENSO on the incidence of dengue infections on five climatological diverse municipalities of Costa Rica, from 2007-2017. Using a GAM and RF approach, we used the weekly climate and dengue cases information from 2007-2016 as a training set, which later allowed us, by using the observed climatological conditions, to predict the dengue cases dynamics of 2017, year that was used as a testing period. 


The article is organized as follows: In Section 2, we provide details on the data and statistical methodology applied to estimate parameters, as well as the description of the model used. In Section 3, we provide the results obtained with the statistical analysis and, in Section 4, we discuss and give our conclusions. 

\section{Materials and Methods} 
 
\subsection{Study areas}
Costa Rica is a tropical country located in the Central American isthmus, between Nicaragua (north), Panam\'a (southeast),the Caribbean Sea (east) and the Pacific Ocean (west), administratively divided into seven provinces and 82 municipalities. With 51,100 square kilometers of land surface, the geographical location of the mountainous system, together with the trade winds, provides numerous and varied micro-climates, dividing the country into seven climatic regions: Central Valley, North Pacific, Central Pacific, South Pacific, North Caribbean, South Caribbean and North Zone, each one further divided into sub-regions \cite{manso2005,imn1}. These multiple micro-climates have played an essential role in shaping the demographic and economic activities of the different regions, providing each one with unique characteristics \cite{minsvivenda}. 

\begin{figure}[h]
\centering
\includegraphics[width=1\textwidth]{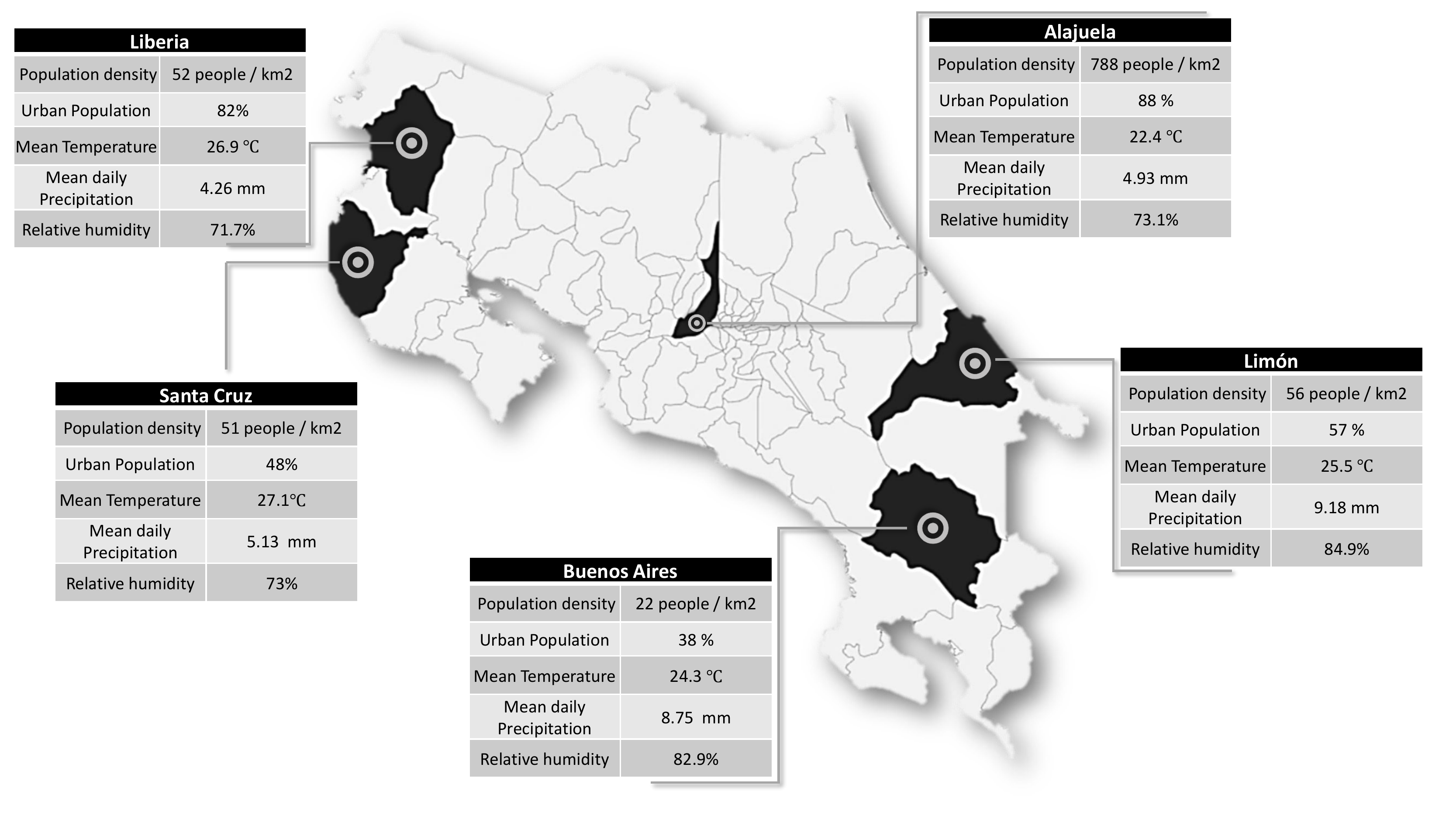}\hfill
\caption{Geographic location of the study areas. Each region has its own climatic patterns and demographic conditions. The mean temperature, relative humidity and precipitation represent the statistical daily average from 2007-2017.}
\label{fig:mapCR}
\end{figure}

Given the climatological diversity, this study was conducted in five municipalities: Santa Cruz and Liberia in the North Pacific, Buenos Aires in the South Pacific, Alajuela in the central part of the country and Lim\'on in the Caribbean coast. Each one with different micro-climates and endemic circulation of the DENV (see Figure \ref{fig:mapCR}). 

Liberia and Santa Cruz are located in the North Pacific climatic region, characterized by being one of the driest and warmest of the country \cite{imn1}. During El Ni\~no years, both Liberia and Santa Cruz, are prone to very extensive dry seasons and droughts, with high economic repercussions to the province \cite{indersc}. After the re-emergence of the \textit{Ae. aegypti} mosquito in Costa Rica, in the 1970s, Liberia was one of the first localities where the vector was identified \cite{morice2010dengue}, it was also, the second municipality to report dengue infections in 1993 and the first to have a case of severe dengue in 1995 \cite{morice2010dengue}. From 2007-2017, Liberia reported a total of 6,685 dengue suspected and confirmed cases,  while Santa Cruz had a total of 10,527 dengue cases \cite{minsa}. Peaks of dengue transmission usually start at the end of May, coinciding with the beginning of the rainy season. 

Buenos Aires is located in the Province of Puntarenas in the South Pacific climatic region. The climate in this municipality is characterized for being rainy with monsoon influence \cite{imn1}. Despite having the adequate conditions for dengue transmission, dengue virus reached the region until 2005 \cite{minsa}. From 2007-2017 a total of 4,405 cases were reported by the Ministry of Health \cite{minsa}, where peaks of transmission vary widely. In the Caribbean coast, Lim\'on, has a decrease in precipitation during the months of March, September and October  \cite{minae}. A total of 7,738 cases were reported during the study period \cite{minsa}. 

Alajuela is the most urban of the study areas. As part of the Central climatic region, this municipality is characterized by a mountainous tropical climate. The Pacific influence in Alajuela, makes this a dry region, making it one of the municipalities of the province where it rains the less \cite{indera}. During the study period a total of 15,158 dengue cases were reported in Alajuela. 

\subsection{Data}
We use two different information sources as main components in the modeling process: observed number of weekly dengue cases and climatological data. 

\subsubsection{Dengue Data}
\begin{figure}
\centering
\includegraphics[width=1.0\linewidth]{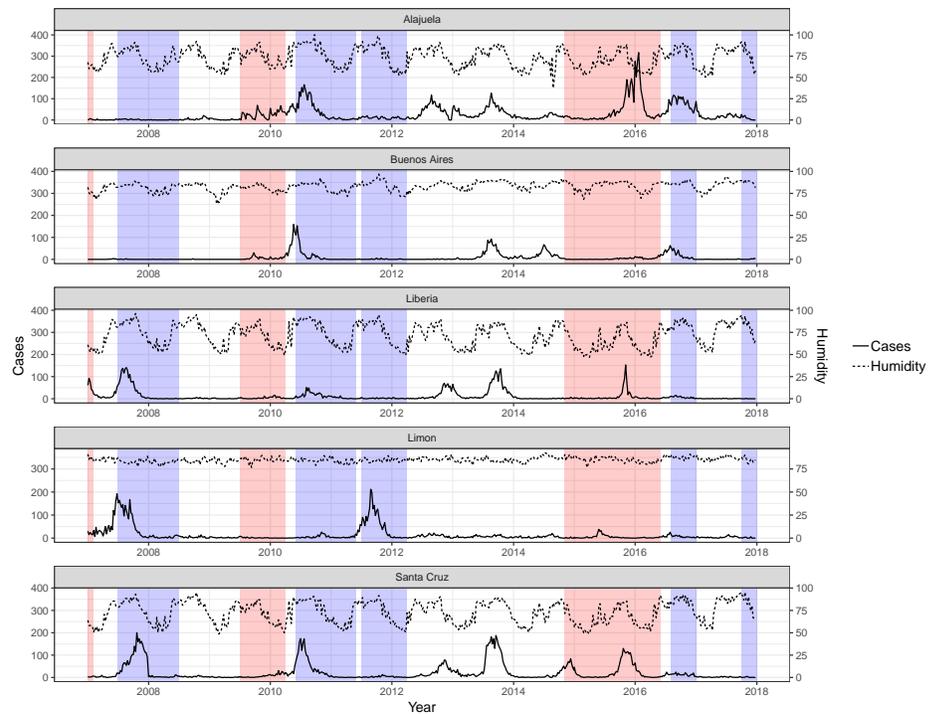}
\caption{Reported Dengue Cases and Relative Humidity in five municipalities of Costa Rica, 2007-2017. During the study period, the municipalities showed variations both in the time and severity of the dengue outbreaks. This seasonality has been linked to the effects of the warm (El Ni\~no) and cold (La Ni\~na) phases of El Ni\~no-Southern Oscillation (ENSO) throughout the country  \cite{ramirez2017,fuller2009}. The cold phase period of ENSO (La Ni\~na) is presented in blue stripes, while the warm phase (El Ni\~no) is presented in red stripes.} 
\label{fig:enso}
\end{figure}

Data on weekly clinically suspected and confirmed dengue cases from Santa Cruz, Liberia, Lim\'on, Alajuela and Buenos Aires, covering the period from 2007-2017 was provided by the Ministry of Health of Costa Rica. In the country, dengue is a mandatory notifiable disease, where both confirmed and probable cases are notified to the Health Surveillance Department from the Ministry of Health  \cite{minsadengue}. Confirmatory diagnosis is made to those patients that live in areas where previous cases and/or confirmed circulation of the dengue virus has not been reported\cite{minsadengue}. Figure \ref{fig:enso}, shows the number of reported dengue cases in La Ni\~na (blue stripe) and El Ni\~no (red stripe) phases from 2007-2017, as well as, the relative humidity during that period.

\subsubsection{Climate data}
Local meteorological data from January 2007- December 2017 was provided by the National Meteorological Institute (IMN) of Costa Rica. A total of five weather stations located in the study areas were active during the eleven-year period: Santa Cruz (40 m a.s.l.), Aeropuerto Liberia Oeste (89 m a.s.l.), Aeropuerto Juan Santamar\'ia in Alajuela (913 m a.s.l.), Aeropuerto Lim\'on (5m a.s.l.) and Pindeco in Buenos Aires (397 m a.s.l.). These weather stations registered daily information of: 

\begin{itemize}
    \item Minimum, Mean and Maximum Temperature: as one of the most important abiotic environmental factors affecting the biology of mosquitoes \cite{alto2013temperature}, the air temperature is defined as "the temperature indicated by a thermometer exposed to the air in a place sheltered from direct solar radiation" \cite{wmotemp} measured in $^{\circ}$C. We will denote the mean temperature as $T$, and we used only this variable due to the large observed correlation among the minimum, mean and maximum over all the study areas.
    \item Precipitation ($P$): is defined as the amount of water that has fallen at a given point over a specified period, expressed in millimeters (mm) \cite{ams}. 
    \item Relative humidity ($RH$) expressed as a percentage (\%), is the ratio of the actual water vapor pressure to the saturation vapor pressure with respect to water at the same temperature and pressure \cite{wmotemp}. 
    \item Weekly ENSO Sea Surface Temperature ($SST$) data was obtained from the Climate Prediction Center (CPC) of the NOAA. After the sea surface temperature was recognized as a key variable in ENSO \cite{rasmusson1982variations}, four regions across the Pacific equatorial belt were defined for measurements (Ni\~no 1+2, Ni\~no 3, Ni\~no 3.4 and Ni\~no 4) \cite{noaa}. We included the SSTA (sea surface temperature anomalies) in the Ni\~no 3.4 region.  
\end{itemize}

Given that all the weather stations had missing observations, we used the method described by  Alfaro and Soley (2009) in \cite{Alfaro2009} and its corresponding implementation in Scilab software v.5.5.2, initially developed by the Institut Nationale de Recherche en Informatique et en Automatique (INRIA). The data was later re-organized to reflect weekly information to match the temporal aggregation of dengue cases data provided by the Ministry of Health. The variable precipitation received a log-transformation to reduce the effect of outlier values, and a constant was added to define the zero cases. 

\subsection{Model Structure and Methods}

The dependent variable that we used along the article is the relative risk of the $i$-th area with respect to the country:

\begin{align*}
  RR_{i,t}=\frac{\frac{\text{Cases}_{i,t}}{\text{Population}_{i,t}}}{\frac{\text{Cases}_{CR,t}}{\text{Population}_{CR,t}}}
\end{align*}
where everything is computed at week $t$ and it is understood as a measure
of relative incidence for the $i$-th study area. In evaluating the effects of climate variables over the
incidence of vector-borne diseases, such as dengue, predictive models such as
Generalized Additive Models and Random Forests have been widely used
\cite{xiao2018weather,carvajal2018machine,liclimate,cabrera}. In what follows we briefly describe
both methods and how the lag information was chosen.

\subsubsection{Choice of covariate lags} \label{sec:lags}
The overall model fit can be improved by adding lagged versions of the covariates. In this way the models can include further information from the past behavior of the variables. Following the ideas of
\cite{hii2012forecast} and \cite{carvajal2018machine}, we determined the largest cross-correlation among the observed cases and each covariate and extracted its respective lag. The maximum allowed lag was taken as 30 weeks. The results are shown in Table \ref{tab:lags} and they are used as input for the models in the next sections.
\begin{table}[H]
  \centering
   \caption{Lags selected by the cross-correlation criteria.}
  \begin{tabular}{lrrrrr}
  \toprule
 & Santa Cruz & Liberia & Lim\'on & Buenos Aires & Alajuela \\ 
  \midrule
Humidity & 5 & 6 & 7 & 7 & 10 \\ 
log(precipitation) & 7 & 7 & 17 & 3 & 10 \\ 
Mean Temp. & 29 & 27 & 4 & 19 & 25 \\ 
SSTA & 27 & 28 & 14 & 27 & 22 \\ 
   \bottomrule
\end{tabular}
 
  \label{tab:lags}
\end{table}

\subsubsection{Generalized Additive Models}\label{sec:GAMs}
A generalized additive model (GAM model) is a generalized linear model defined as a linear combination of smooth functions of covariates \cite{Wood2017}. Its main advantage is the flexibility on the specification of the relationship between a dependent variable and its corresponding covariates, contrary to the classical way to model
that relationship based on linear associations, which is not always a good assumption in many applications. The general form of a GAM model is:
\begin{align}\label{eq:1}
  g(\mu_i)=\sum_{j=1}^Kf_j(x_{ij})
\end{align}
where $\{Y_i\}$ is an independent sample of observations with their respective means $\{\mu_i\}$ and distributed as a member of the exponential family \cite{Hastie1987}. The K covariates $x_{.j}$ are evaluated on the smooth
functions $f_j$ and the terms in equation \eqref{eq:1} can also contain interactions between covariates. The functions $f_j$ are chosen in most cases as penalized regression splines \cite{Wood2017}. Penalized likelihood estimation is employed to fit the parameters in GAM models \cite{OSullivan1986}.

For our purposes we defined the GAM model for a single study unit as follows:
\begin{align}\label{eq:2}
  RR_t=f_1(RR_{t-1})+f_2(RH_t)+f_3(RH_{t-l_1})+f_4(\tilde P_t)+f_5(\tilde P_{t-l_2})+\\ \nonumber f_6(T_t)+f_7(T_{t-l_3})+f_8(SST_t)+f_9(SST_{t-l_4})+\epsilon
\end{align}
where we remove the subscript $i$ for convenience, the covariate $\tilde P_t:=\log (P_t)$, the lags $\{l_1,l_2,l_3,l_4\}$ are taken from Table \ref{tab:lags}, $\epsilon$ is a Gaussian error and the smooth functions are penalized cubic regression splines. The estimation process of the GAM model was performed with the R package \verb|mgcv| \cite{Wood2011}. 

\subsubsection{Random Forests}\label{sec:RFs}
The essential idea of Random Forest is to construct an ensemble of trees based on bootstrapping techniques and the predicted values are computed using averages over the ensemble to reduce the excess of prediction variance
\cite{Breiman2001,Hastie2009}. This technique has several advantages over other boosting methods, the prediction accuracy is attained by including sequentially the covariates in order to maximize the efficiency of each tree. Besides, the computational manipulation in terms of parameter tuning is not existent \cite{Hastie2009}.     

For this application we used the same set of covariates and dependent variables as in equation \eqref{eq:2}. The training and prediction process was done with the R packages \verb|caret| and \verb|randomForest| \cite{caret,RF} with approximately 500 sample trees.

\section{Results}
Based on the number of dengue reported cases and weather information from 2007-2017, we fitted the prediction models described in sections \ref{sec:GAMs} and \ref{sec:RFs}. We took the weekly information of both the dependent variable and covariates over the period 2007-2016 as a training set for both methods and the 52 weeks of 2017 as a testing period. Both methods were also fitted using the number of weekly observed cases as a dependent variable, but we prefer to show the models fitted with the relative risk due to ease of comparison among study areas.

\begin{figure}
  \centering
  \includegraphics[scale=0.5]{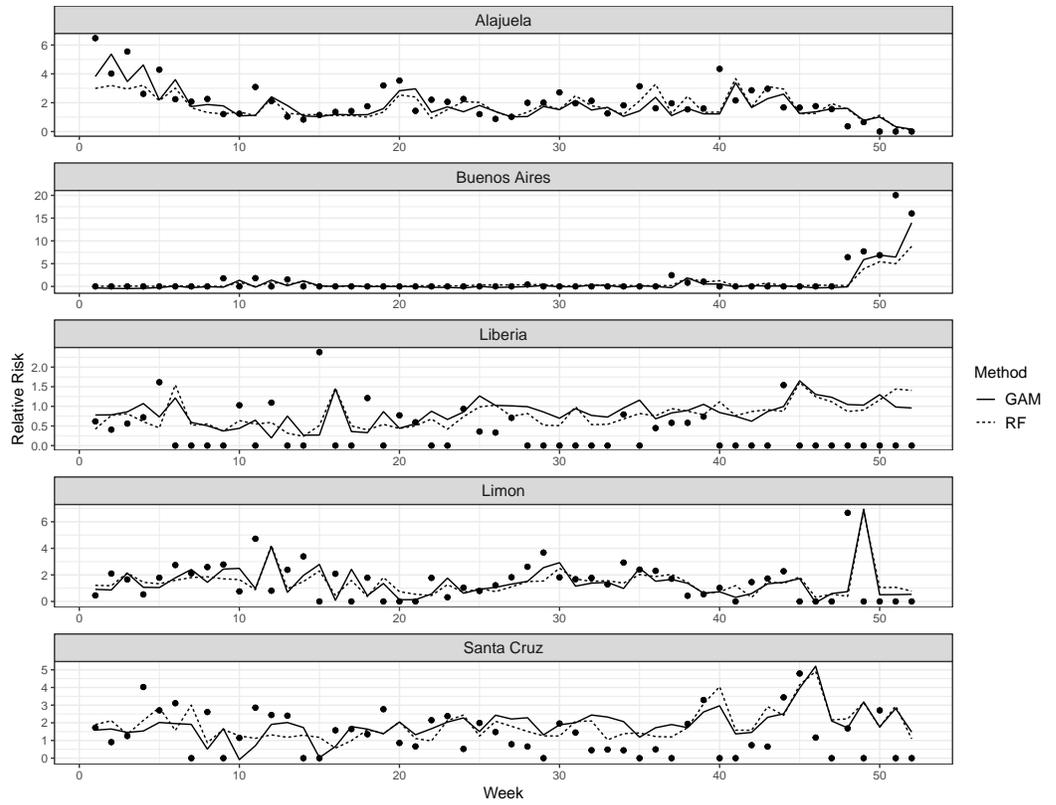}
  \caption{Model comparison over the prediction period. Lines: predicted relative risk. Points: observed relative risk.}
  \label{fig:comparison}
\end{figure}

Figure \ref{fig:comparison} shows the results of the two different statistical models used to predict the incidence of dengue in 2017. The dotted and solid lines, correspond to the predicted relative risk of each study area over the testing period.

The predicted RR of Alajuela is quite impressive because it recovers the general decreasing trend in the observed behavior of the series and it also can capture weeks where the incidence increases suddenly. It is also interesting to note the Lim\'on and Buenos Aires areas, where along 2017 there were some peaks of transmission, and the model was able to predict successfully the general behavior of those events within one week of precision. Santa Cruz and Liberia (both located on the Pacific Coast) were the areas with more difficulties in terms of prediction, but
we still were able to predict weeks with an increasing or decreasing incidence precisely. These two study areas are particularly marked with seasonal effects that can increase the serial variance within the testing period, and hence the prediction does not perform as well. 

Table \ref{tab:RMSE} contains the Normalized Root-Mean-Square Error (NRMSE) of each combination of method and study area.
\begin{table}
  \centering
   \caption{Normalized Root-Mean-Square Error among prediction methods.}
  \begin{tabular}{lrr}
  \toprule
 & GAM & RF \\ 
  \midrule
Alajuela & 0.53 & 0.54 \\ 
  BuenosAires & 1.74 & 2.03 \\ 
  Liberia & 2.43 & 2.36 \\ 
  Limon & 1.21 & 1.24 \\ 
  SantaCruz & 1.13 & 1.14 \\ 
   \bottomrule
\end{tabular}
  \label{tab:RMSE}
\end{table}
The NRMSE is defined as follows:
\begin{align*}
  NRMSE=\frac{1}{\overline{RR_{obs}}}\sqrt{\frac{1}{52}\sum_{i=1}^{52}(\widehat{RR}_i-RR_{i,obs})^2}
\end{align*}
where $\widehat{RR}_i$ is the predicted relative risk and $RR_{i,obs}$ is the observed relative risk at week $i$. $\overline{RR_{obs}}$ is the mean of the observed relative risk over the testing period. We used this measure to compare the attained dispersion of the prediction  with respect to its mean behavior. Note that the best prediction in terms of this measure is attained in Alajuela followed by Buenos Aires and Santa Cruz, which is relatively consistent to the conclusions of Figure \ref{fig:comparison}. 

\section{Discussion}
With the recent emergence of chikungunya and Zika, into the country, as well as, the continuous high incidence of dengue infections \cite{minsa}, the burden of \textit{Aedes} transmitted diseases has significantly increased. In a country where resources for vector control are limited, the urgency to implement effective and affordable vector control mechanisms to complement existing ones \cite{world2012global} is at the forefront of public health policy in Costa Rica.

As the transmission dynamics of dengue infections are inextricably linked to the interplay of multiple meteorological conditions, recently significant advances in climate data availability, statistical modeling and information technology \cite{worldclimate}, has increasingly opened the possibility of using climate information as effective predictors of dengue incidence \cite{halide2008predictive,descloux2012climate,lowe2017climate}. However, in Costa Rica, a country with tropical conditions optimal for mosquito survival, the extent of influence that different climate variables have on local dengue epidemiology, and the possibility of using them as early warning signals, is still in its early stages. Although different studies have been conducted \cite{troyo2008,fuller2009,ramirez2017}, the presence of multiple micro-climates, separated by short distances, makes it relevant to advocate for more localized analyses that can take into account the specific and unique characteristics of each municipality.  


In the current study, we collected weekly dengue incidence provided by the Ministry of Health, observed local temperature, precipitation and humidity from five different weather stations provided by the National Meteorological Institute, and SSTA information from 2007-2017 that could allow us to test the predictive capacity of the two selected models, Random Forest and Generalize Additive models, as well as, the level of climatological influence in the epidemiology of dengue infections in the selected municipalities. 

Our analyses showed that while using the 2007-2016 period as a training set, both, the Generalized Additive Models and Random Forest performed well in predicting the temporal patterns of dengue incidence in 2017, a year that was used as a testing period. The results demonstrated that, even when the number of cases were low, as it was the case in Buenos Aires, the model accurately predicted the onset of the outbreak. However, its predictive accuracy differed depending on each region, as localities in the North Pacific coast, Liberia and Santa Cruz, the model overpredicted the number of cases. Hence, further exploration is needed to identify if in fact the model overpredicted the number of cases or there was under-reporting by the health officials in those specific regions. In a disease with such diverse and unspecific symptoms, during 2017 the laboratory responsible for coordinating the virological surveillance of arbovirus at a national level, highlighted in their annual report the low number of samples sent for dengue confirmation by municipalities in the province of Guanacaste during that year, identification that is crucial to monitor the behavior of the virus \cite{inciensa2018}. Also, other factors intrinsic to the local epidemiological dynamics are likely to play a crucial and different role for certain years among the different locations. Variables such as socio-economic conditions, human-mobility, population herd immunity for different dengue serotypes, the intensity of public health strategies, where increased control activities during certain periods of the year, such as, the beginning of the school year in Mexico \cite{Hernandez2016}, can significantly change the dynamics of dengue transmission, were not included in the model, therefore limiting the accuracy of prediction.


The efficacy of the model also depends on the availability of accurate climate information over the training and testing periods. In its current form, the model uses observed climatological conditions as covariate variables, limiting the prediction process on the availability of such information over the study areas. In addition, all of the weather stations presented missing information, therefore a statistical method was used to complete the series. The development of accurate climate forecasts represents a major challenge, particularly due to the low timescales in the forecasting methods of the country. Further work is in progress to explore alternative sources of local meteorological information as predictors of DENV incidence.

Despite these limitations, results from this study, suggest that large-scale climate and local weather factors can potentially be used as effective tools in the decision-making process of local public health-authorities. It also shows, as in previous studies \cite{fuller2009,sanchez2019}, the importance of statistical models as instruments in the rapid analysis of information generated by different local and national institutions, as they could enhance the management of early epidemic response and preventive measures in Costa Rica. However, the development of tailored climate products and services that can be fully mainstreamed into public health decision-making, is a collaborative process that would require inter-institutional integration of expertise and data \cite{wmorisk}, including the Ministry of Health, the National Meteorological Institute and the National Census Bureau, among others, collaboration that could have a positive impact in the management not only of mosquito-borne diseases, but all the other climate-sensitive diseases that affect the country. 

\section*{Acknowledgements}
We thank the Research Center in Pure and Applied Mathematics and the Mathematics Department at Universidad de Costa Rica for their support during the preparation of this manuscript. The authors gratefully acknowledge institutional support for project B8747 from an UCREA grant from the Vice Rectory for Research at Universidad de Costa Rica. We would like to thank the Ministry of Health and the National Institute of Meteorology for providing the necessary dengue incidence data and climate information. We also thank Oscar Calvo-Solano, for his help in completing the climate data. This article is part of a thesis project for the masters in Public Health at the University of Costa Rica.    


\end{document}